\documentclass{article} 
\usepackage{arxiv_submission,times}
\usepackage{hyperref}
\usepackage{url}
\usepackage{graphicx}
\usepackage[title, titletoc]{appendix}
\usepackage{multirow}
\usepackage{hyperref}

\title{Robustness of Rotation-Equivariant \\ Networks to Adversarial Perturbations}

\author{B\'eranger Dumont, Simona Maggio \& Pablo Montalvo \thanks{PM is currently Visiting Scientist at the Rakuten Institute of Technology Paris.}\ \ \thanks{All three authors contribued equally to this work.} \\
Rakuten Institute of Technology Paris \\
\texttt{\{beranger.dumont,simona.maggio\}@rakuten.com}, \\
\texttt{pablo.montalvo.leroux@gmail.com} \\
}

\begin{document}

\maketitle

\begin{abstract}
Deep neural networks have been shown to be vulnerable to adversarial examples: very small perturbations of the input having a dramatic impact on the predictions.
A wealth of adversarial attacks and distance metrics to quantify the similarity between natural and adversarial images have been proposed, recently enlarging the scope of adversarial examples with geometric transformations beyond pixel-wise attacks.
In this context, we investigate the robustness to adversarial attacks of new Convolutional Neural Network architectures providing equivariance to rotations.
We found that rotation-equivariant networks are significantly less vulnerable to geometric-based attacks than regular networks on the MNIST, CIFAR-10, and ImageNet datasets.

\end{abstract}

\section{Introduction}
Deep learning provided significant breakthroughs in machine learning, and Convolutional Neural Networks (CNNs) are now being used routinely for computer vision tasks. However, theoretical as well as practical concerns remain, a prime example being adversarial attacks: very small perturbations of the input causing catastrophic changes in the predictions of the network.
Since adversarial examples were first noticed \citep{Szegedy13,Goodfellow14}, the vast majority of the studies has focused on pixel-wise attacks for the $L_0$, $L_2$, or $L_{\infty}$ distance metrics on images (for a review, see, {\it e.g.}, \citet{advreview}).
Very recently the notion of adversarial examples on images was refined to account for the human perception \citep{imperceptible}, and extended to rigid geometric transformations \citep{manitest,epflpaper,mitpaper} and local geometric distortions \citep{stnflow}.

In an independent effort, several groups strived to extend the symmetry properties of CNNs beyond equivariance to translations. The most natural next step is to equip the models with invariance or equivariance to rotations, so that a rotation of the input image will either leave the feature maps unchanged or will rotate them accordingly. Most of the existing approaches fall into two categories: the ones based on the rotation of the input images \citep{stn,cyclic,tipooling,warpedconv,polartrans}, and the ones based on constraints on the structure of the filters \citep{cohen16,steerablecnns,hnet,orn,vectorfieldnetworks,deeprot,steerablefilters}.

We assess the robustness to adversarial examples of three distinct rotation-equivariant CNN architectures: Group Equivariant Convolutional Neural Networks (G-CNNs, \citet{cohen16}), Harmonic Networks (H-Nets, \citet{hnet}), and Oriented Response Networks (ORNs, \citet{orn}). First, G-CNNs provide equivariance to 90 degrees rotations and mirror reflections by redefining the convolution operator over symmetry groups. Second, H-Nets obtain equivariance to rotations of arbitrary angles using complex-valued filters constrained to the family of circular harmonics. Finally, ORNs obtain invariance to rotations using filters that actively rotate during convolution.

All of these models have achieved (or defined a new) state-of-the-art performance on the rotated MNIST dataset (MNIST-rot, \citet{rotmnist}, while G-CNNs and ORNs have proven to be competitive on the CIFAR-10 dataset \citep{cifar10} as well. All networks provide patch-wise rotation equivariance, thus might be more robust to local geometric transformations.
Deep Rotation Equivariant Networks \citep{deeprot} are also competitive on MNIST-rot and CIFAR-10, but must be equivalent to G-CNNs with the group $p4$, up to differences in the proposed architecture and learning procedure \citep{generalequiv}.
We present our experimental setup in Section~\ref{expsetup}, our results in Section~\ref{results}, and we conclude in Section~\ref{conclusions}.

\section{Experimental setup}
\label{expsetup}

We train rotation-equivariant CNNs, as well as regular CNNs for comparison, on the MNIST, CIFAR-10 and ImageNet \citep{imagenet} datasets. We then evaluate those models against adversarial examples generated from the corresponding datasets. 
In the case of MNIST, we consider models with seven $5\times5$ convolutional layers, following the definition from \citet{cohen16,hnet}. We suppress as much as possible architecture-related specificities to compare the models on equal grounds, and consider the same number of parameters of about $34$k for all models.

As for CIFAR-10, we consider a 44-layer residual network, ResNet-44, and its corresponding G-CNN version which we call G-ResNet. We follow the exact same procedure as \citet{cohen16}, and adjust the number of filters per layer in order to obtain the same number of parameters for both models ($2.6$M). We were not able to achieve a competitive accuracy with H-Nets.

In the case of ImageNet, to the best of our knowledge only ORNs have public results. A pre-trained residual model with 18 convolutional layers, OR-ResNet-18, was provided by the authors. We compare its performances to the standard Torch implementation of ResNet-18. Both models have a total of $1.4$M parameters. For the training and validation of all networks, we have used publicly available implementations.\footnote{
See \href{https://github.com/tscohen/GrouPy}{\tt tscohen/GrouPy},
\href{https://github.com/tscohen/gconv_experiments}{\tt tscohen/gconv\_experiments} (G-CNN), \\
\href{https://github.com/deworrall92/harmonicConvolutions}{\tt deworrall92/harmonicConvolutions} (H-Net),
\href{https://github.com/ZhouYanzhao/ORN}{\tt ZhouYanzhao/ORN} (ORN), and \\
\href{https://github.com/facebook/fb.resnet.torch}{\tt facebook/fb.resnet.torch} (ResNet baseline on ImageNet) on \href{https://github.com/}{\tt GitHub.com}.
}

We consider recently proposed adversarial attacks based on geometric transformations. First, rigid geometric transformations (global rotations and translations), following closely the procedure of \citet{mitpaper}, but considering a different range: translations of $\pm 3$ pixels on both axes, and rotation of $\pm10$ degrees by step of one degree for all datasets. In the case of CIFAR-10, we have compared zero and edge paddings, and found no significant difference. For ImageNet, we perform rotation and translation of the $256 \times 256$ images before cropping to $224 \times 224$.

Next, we consider spatially transformed adversarial examples (stAdv, \citet{stnflow}), which are white-box targeted attacks based on Spatial Transformer Networks \citep{stn}.
The generated adversarial examples are designed to be misclassified while keeping the spatial transformation distance low. To balance adversarial and flow losses, as defined in \citet{stnflow}, we take $\tau=0.10$.
For each sample we generate a set of stAdv attacks, taking each possible wrong label as a target.
As there was no publicly available implementation of stAdv, we have implemented them in TensorFlow.\footnote{Our implementation is publicly available at \href{https://github.com/rakutentech/stAdv}{\tt rakutentech/stAdv} on \href{https://github.com/}{\tt GitHub.com}.}

We also wanted to assess the robustness of rotation-equivariant networks to popular pixel-wise attacks on the $L_p$ norm. To that end, we considered the Fast Gradient Sign \citep{Goodfellow14} and DeepFool \citep{deepfool} attacks. They managed to completely fool the classifiers in almost all our cases, showing that there is no significant added robustness to these attacks from rotation-equivariant architectures.

\section{Results}
\label{results}

For every model and dataset, we report results on the single-crop error rate on the natural test set ({\it i.e.}, with no adversarial perturbation).
The robustness to adversarial attacks is quantified with the \textit{attack success rate} (ASR): the average fraction of attacks fooling a classifier, for a given type of attack. In the computation of the ASR, we exclude samples from the test set which are misclassified even if no perturbation is applied.

For the models trained on MNIST, we obtained a test error of less than $0.7\%$ for all cases. The results are shown in Table~\ref{Full-table} (top). We find that H-Nets perform worse than the baseline on rotation-based attacks, which is consistent with the observations of \citet{deeprot}. However, they tend to be more accurate under spatially transformed adversarial attacks, which could show that H-nets are robust to local deformations due to their equivariance to local rotations of arbitrary angles. G-CNNs are found to be more robust to rigid geometric transformations than the CNN baseline, which shows that the learned representations are useful against this type of attack.

\begin{table}[t]
\caption{Error on the natural test set and attack success rate (ASR) on the MNIST (top), CIFAR-10 (bottom left), and ImageNet (bottom right) datasets. R+T indicates that the attacks are made from a combination of rotations and translations, while R and T are for rotation- and translation-only attacks, respectively. stAdv are spatially transformed adversarial attacks.}
\label{Full-table}
\begin{center}
\begin{tabular}{c c c c c}
{\bf MNIST} & & {\bf CNN} & {\bf H-Net} & {\bf G-CNN}
\\ \hline \\
\multicolumn{1}{c}{Error} & & 0.68\% & 0.69\% & 0.55\% \\
\hline
\multirow{4}{*}{ASR} & R+T       &   6.2\%     &    20.5\%           &    \textbf{3.4\%} \\
& R        &   1.2\%     &    4.8\%            &    \textbf{0.9\%} \\
& T      &   2.6\%     &    17.6\%           &    \textbf{1.4\%} \\
& stAdv           &   92.6\%    &    \textbf{77.7\%}  &    80.1\% \\
\end{tabular}
\end{center}

\begin{minipage}{0.5\textwidth}
\begin{center}
\begin{tabular}{c c c c c c}
\\
{\bf CIFAR-10} & & {\bf ResNet} & {\bf G-ResNet}
\\ \hline \\
Error & & 8.9\%  & 6.1\% \\
\hline
\multirow{4}{*}{ASR} & R+T     &   88.4\%        &      \textbf{70.1\%}     \\
 & R      &   77.9\%        &      \textbf{52.1\%}     \\
 & T    &   88.8\%        &      \textbf{66.5\%}     \\
 & stAdv         &   83.4\%        &      \textbf{83.1\%}     \\
\end{tabular}
\end{center}
\end{minipage}
\begin{minipage}{0.5\textwidth}

\begin{center}
\begin{tabular}{c c c c c c}
{\bf ImageNet} & & {\bf ResNet} & {\bf OR-ResNet}
\\ \hline \\
Error & & 30.6\%  & 28.9\% \\
\hline
\multirow{3}{*}{ASR} & R+T &    6.8\%           &   \textbf{5.7\%}   \\
 & R  &    6.5\%           &   \textbf{5.4\%}   \\
 & T  &    \textbf{3.4\%}  &   4.3\%            \\
\end{tabular}
\end{center}

\end{minipage}
\vspace{-2mm}
\end{table}

Second, on the CIFAR-10 dataset, we trained a ResNet-44 and obtained a test error of $8.9\%$, and trained a rotation-equivariant G-ResNet and obtained a test error of $6.1\%$. Results are shown in Table~\ref{Full-table} (bottom left).
All attacks are much more successful than on MNIST since CIFAR-10 images have a much richer content while having a comparable size.
Overall, the G-ResNet achieves far lesser vulnerability to attacks by rotation and translation than the regular ResNet, and is only marginally better against the stAdv attacks. For pure rotations, we have a drop in the ASR of $33\%$.

Finally, on the larger ImageNet dataset, we used a pre-trained ResNet-18 (resp.\ OR-ResNet-18) that achieves $30.6\%$ error (resp.\ $28.9\%$) on the test set.
Results are shown in Table~\ref{Full-table} (bottom right). We can see that ORNs bring an improvement against rotation attacks, and are slightly worse than a regular ResNet with translation attacks. For combined rotations and translations, there is a drop in the ASR of $19\%$.
We only tested rotation and translation-based attacks and could not test the stAdv attacks for the OR-ResNet given the absence of a Torch implementation for this kind of attack.

We also analyzed the distribution of the ASR for combined rotations and translations on the three datasets. Results are shown in Fig.~1. In the case of MNIST, the CNN and G-CNN distributions are smoothly falling and very similar to each other, while the one for H-Net exhibits a more complicated structure with a secondary peak around ${\rm ASR} = 0.3$.
As for CIFAR-10, both ResNet and G-ResNet produce clear bimodal distributions (less pronounced in the G-ResNet case). We have found that all samples from the ``bird'' class have $0 < {\rm ASR} < 0.2$ for the regular ResNet, indicating that the network has learned discriminative features which are nearly invariant to the considered transformations for this class alone.
Finally, in the case of ImageNet, both distributions are smoothly falling but at a much lower rate than for MNIST.

\begin{figure}[h]
\label{fig_MNIST}
\begin{center}
\begin{minipage}{0.495 \linewidth}
\includegraphics[width = \linewidth]{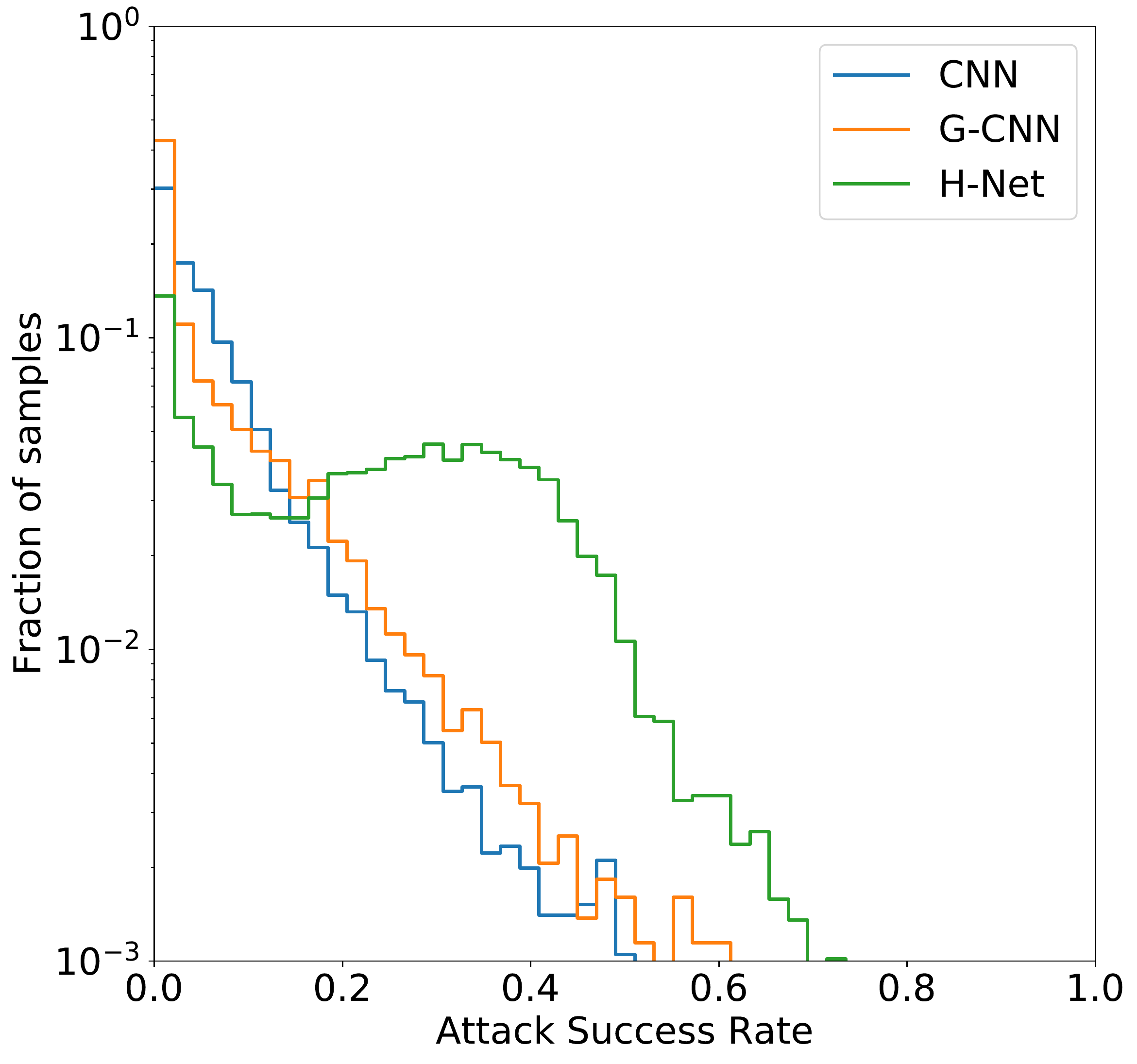}
\end{minipage}
\begin{minipage}{0.495 \linewidth}
\includegraphics[width = \linewidth]{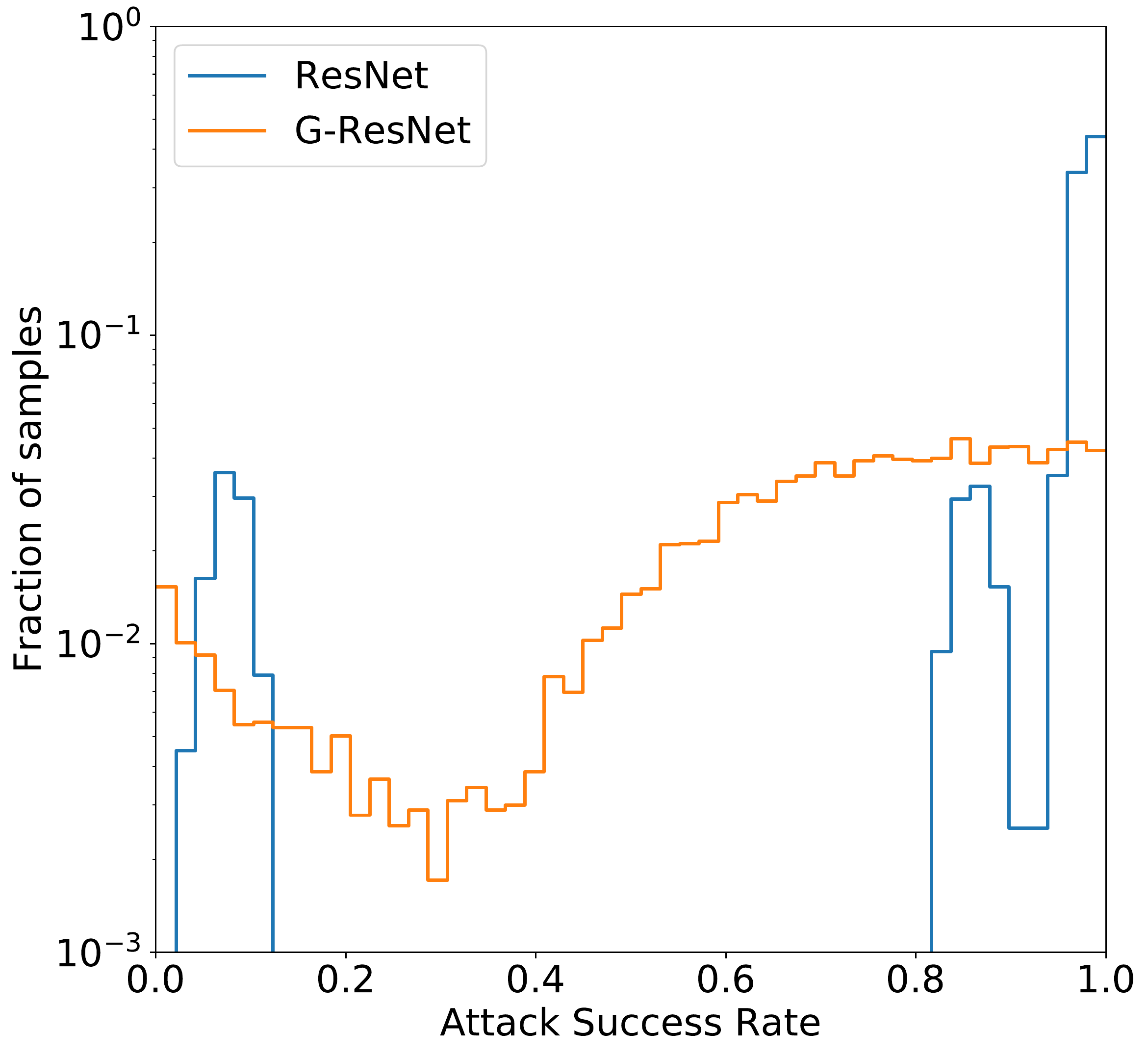}
\end{minipage}
\begin{minipage}{0.495 \linewidth}
\includegraphics[width = \linewidth]{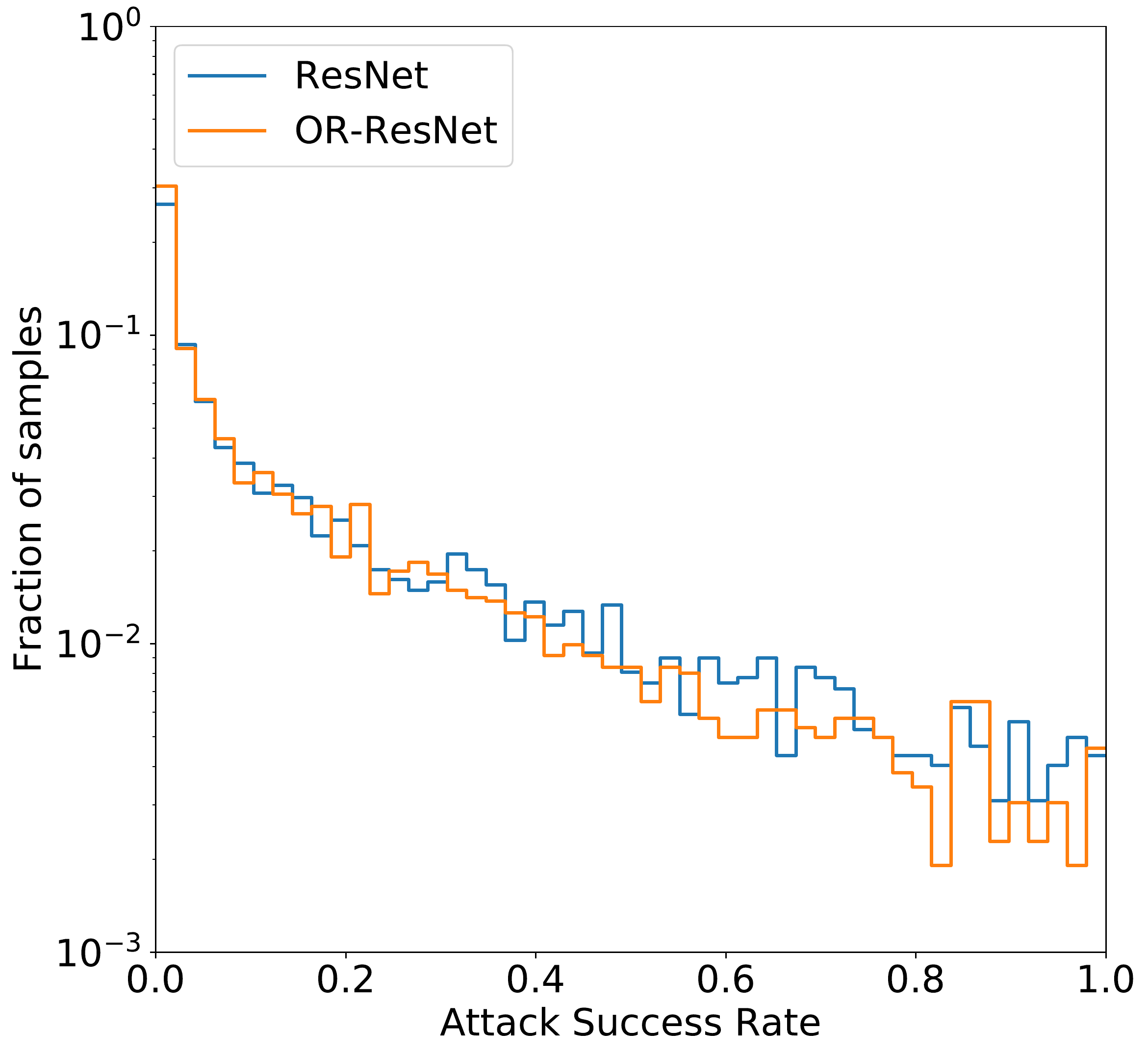}
\end{minipage}
\end{center}
\caption{Distributions of the attack success rate for combined rotations and translations on the test set for MNIST (top left), CIFAR-10 (top right), and ImageNet (bottom).}
\end{figure}

Finally, we checked the correlation coefficient $\rho$ between the ASR and the absolute value of the rotation angle. In the case of MNIST, all tested rotation-equivariant networks are less sensitive to the rotation angle than the regular CNN, with $(\rho_{\rm CNN}, \rho_{\rm H-Net}, \rho_{\rm G-CNN}) = (0.28, 0.12, 0.18)$.
Surprisingly, we observe the opposite behavior in the case of the residual networks we consider for CIFAR-10 ($(\rho_{\rm ResNet}, \rho_{\rm G-ResNet}) = (0.07, 0.33)$), while there is no significant difference for the two models tested on ImageNet ($(\rho_{\rm ResNet}, \rho_{\rm OR-ResNet}) = (0.06, 0.05)$).

\section{Conclusions}
\label{conclusions}

The investigation of changes in the architecture of CNNs as a defense against adversarial examples is a relatively unexplored area.
In order to evaluate the robustness of rotation-equivariant networks to adversarial examples, we conducted experiments on three different types of CNN architectures: G-CNNs, H-Nets, and ORNs, on three different datasets: MNIST, CIFAR-10, and ImageNet. The networks which are equivariant to rotations by discrete angles were found to be significantly more robust to attacks based on small translations and rotations and, more marginally, to attacks based on local geometric distortions.
 We hope that this work will serve as a motivation for further studies on CNNs with extended symmetries, as well as for exploring the interplay between the natural robustness of the rotation-equivariant architectures to geometric-based adversarial attacks and other mechanisms of defense against adversarial examples.

\subsubsection*{Acknowledgments}

We thank Andr\'es Hoyos-Idrobo for the useful discussions and Laurent Ach for his support.

\bibliography{adversarial_roteq}
\bibliographystyle{arxiv_submission}

\end{document}